\documentclass{article}
\usepackage{nips15submit_e}
\usepackage{times, amsmath, bbm, graphicx, amsmath, amssymb, multirow, lscape}
\usepackage{url}
\usepackage[utf8]{inputenc}
\usepackage{pdfpages}

\usepackage[backend=bibtex]{biblatex}
\usepackage[colorlinks]{hyperref}
\hypersetup{
    bookmarks=true,         
    unicode=false,          
    pdftoolbar=true,        
    pdfmenubar=true,        
    pdffitwindow=false,     
    pdfstartview={FitH},    
    pdftitle={Improving the Neural Algorithm of Artistic Style},    
    pdfauthor={Roman Novak and Yaroslav Nikulin},     
    pdfsubject={Improving the Neural Algorithm of Artistic Style},   
    pdfkeywords={Style Transfer, Image Synthesis, Deep Convolutional Neural Networks}, 
    pdfnewwindow=true,      
    linkcolor=red,          
    citecolor=green,        
    filecolor=magenta,      
    urlcolor=cyan           
}

\title{Improving the Neural Algorithm of Artistic Style}

\author{
Roman Novak \textnormal{and} Yaroslav Nikulin\\
Department of Mathematics\\
École normale supérieure de Cachan\\
94230 Cachan, France \\
{\texttt{\{}\href{mailto:rnovak@ens-cachan.fr}{\texttt{rnovak}}\texttt{, }\href{mailto:yaroslav.nikulin@ens-cachan.fr}{\texttt{yaroslav.nikulin}}\}\texttt{@ens-cachan.fr}} \\
}

\begin{document}
\maketitle

\begin{abstract}
	In this work we investigate different avenues of improving the Neural Algorithm of Artistic Style \cite{Gatys15}. 
	
	While showing great results when transferring homogeneous and repetitive patterns, the original style representation often fails to capture more complex properties, like having separate styles of foreground and background. This leads to visual artifacts and undesirable textures appearing in unexpected regions when performing style transfer.
	
	We tackle this issue with a variety of approaches, mostly by modifying the style representation in order for it to capture more information and impose a tighter constraint on the style transfer result.
	
	In our experiments, we subjectively evaluate our best method as producing from barely noticeable to significant improvements in the quality of style transfer.
\end{abstract}

\begin{figure}[t]
		 	\centering
		 	\includegraphics[width=0.195\textwidth]{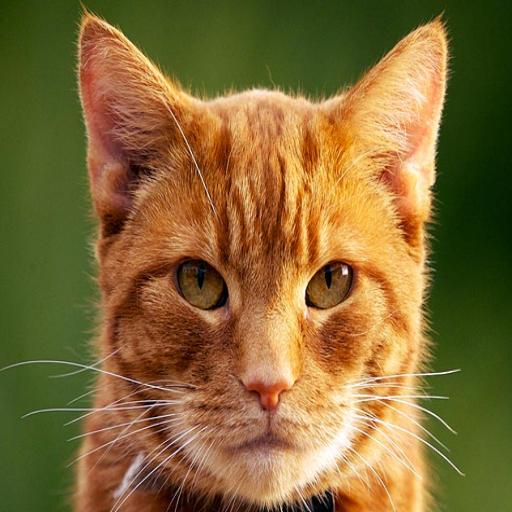}
		 	\includegraphics[width=0.195\textwidth]{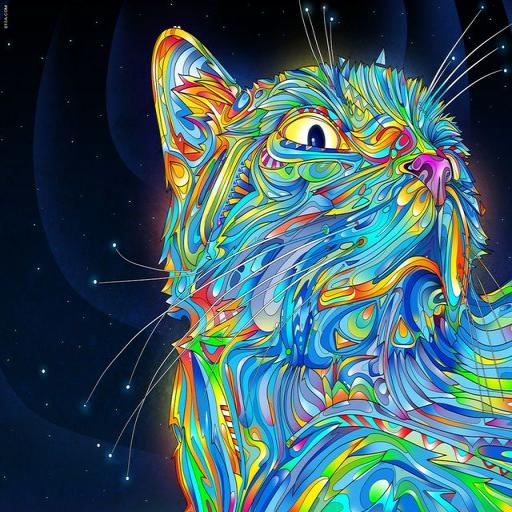}
		 	\includegraphics[width=0.195\textwidth]{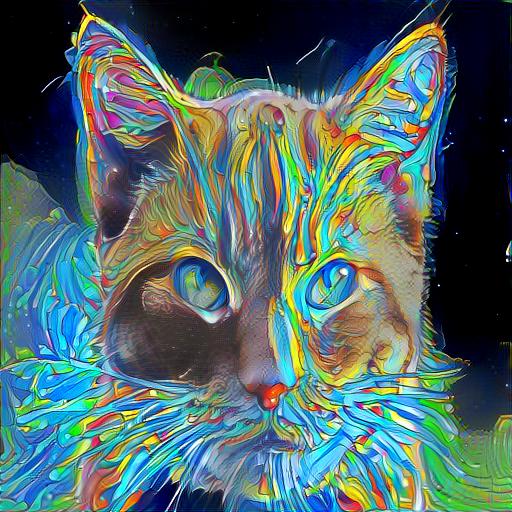}
		 	\includegraphics[width=0.195\textwidth]{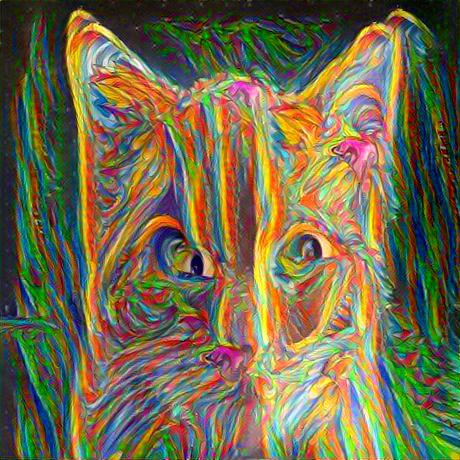}
		 	\includegraphics[width=0.195\textwidth]{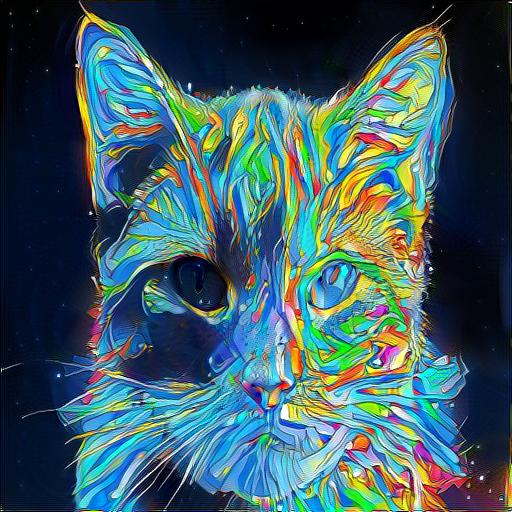}
		 	\caption{Different style transfer methods. Left to right: content  \cite{cat}, style \cite{cat3}, Gatys \textit{et al.}  \cite{Gatys15}, Li and Wand \cite{CNNMRF}, ours.}
		 	\label{front}
\end{figure}

\section{Introduction}\label{intro}
    We first briefly describe the original algorithm of style transfer presented in \cite{Gatys15}. 
    
    In the following we denote $F^l$ the 3D output (activation / feature volume) of a convolutional layer $l$ of size $K_l \times X_l \times Y_l$ with $K_l$ being the number of convolutional filters (features, kernels) in layer $l$, $X_l$ and $Y_l$ being the spatial dimensions. 
    
    Using a pre-trained object recognition deep convolutional neural network the content of an image is defined as an activation volume $F^{l_c}$ of a fixed (usually relatively deep) layer $l_c$ of the network given the image as an input. In \cite{Gatys15} the VGG \cite{VGG} network was used and the output of \texttt{conv4\_2} layer was considered as content. 
    
    Style is defined on top of activation volumes as a set of Gram correlation matrices $\left\{\mathcal{G}_l\right\}$ with the entries $\mathcal{G}^l_{i j} = \langle F^l_i, F^l_j\rangle$, where $l$ is the network layer, $i$ and $j$ are two filters of the layer $l$ and $F^l_k$ is the activation map (indexed by  spatial coordinates) of filter $k$ in layer $l$. In other words, $\mathcal{G}^l_{i j}$ value says how often features $i$ and $j$ in the layer $l$ appear together in the input image. 
    
    Once defined, the content and style spaces can be used to find a synthesized image $I$ matching both given content $C$ and style $S$ images, resulting in high-quality repainting of the content image with preserved semantic information but covered  with textures extracted from the style image. The resulting image $I$ is synthesized by back-propagating the loss
    
    $$
		\alpha \texttt{StyleLoss}(S, I) + \texttt{ContentLoss}(C, I) =\alpha\sum_{l}w_l\left\|\mathcal{G}^l(S)-\mathcal{G}^l(I)\right\|^2 + \left\|F^{l_c}(C) - F^{l_c}(I)\right\|^2    		
    $$
    
    into pixels of $I$ ($\alpha$ is the style/content trade-off hyper-parameter and $w_l$ is a style weight assigned to a specific layer, set to 1 for all layers by default).

    The suggested synthesizing algorithm generally produces great results in transferring repetitive artistic styles. Unfortunately, otherwise generated images often don't meet human expectations of how the style should be transferred. This is the problem we attempt to tackle in this work.
    
    Further, when visually inspecting style transfer results and commenting on the quality, we use two complimentary criteria that we expect a good style transfer algorithm to meet:
    \begin{enumerate}
    	\item Similar areas of the content image should be repainted in a similar way.
    	\item Different areas should be repainted differently.
    \end{enumerate}
     As it turns out, it is often difficult to satisfy both simultaneously. 
     
     The rest of the work is structured as follows: section \ref{rel} gives an overview of other research built on top of the original style transfer algorithm. Section \ref{cm} presents our suggested improvements to the style transfer algorithm. Section \ref{exp} describes and presents the results of experiments conducted with a visual comparison of different approaches to style transfer. We give concluding remarks in section \ref{conc}.

\section{Related Work}\label{rel}
    Several works related to \cite{Gatys15} have been published recently, both introducing new applications of style transfer and improving the algorithm itself. 
    
    In \cite{ContentAware} the impact of resolution and content-style alignment are studied, however, a significant part of suggested pipeline was done manually in photo-editing software and seems to be applicable only to a pair of strongly similar content-style images. Another novelty presented in this paper is super-resolution style transfer when a higher resolution style image is iteratively transferred to a lower resolution content image, upsampled at every iteration.
    
    To improve the style transfer and photorealism of the generated images, a combination of dCNNs and Markov Random Fields is used in \cite{CNNMRF} to leverage  matching feature patches in content and style images. Instead of global bag-of-features-like approach as in \cite{Gatys15}, \cite{CNNMRF} works with best corresponding patches of feature maps: MRF prior gives better local plausibility of the image. Visually,  results produced by \cite{CNNMRF} resemble repainting the content image with appropriate pieces from the style image, while the original algorithm \cite{Gatys15} can be informally described as generating textures of different scale and complexity over the content image. It should be noted that, while giving more photorealistic results (with a photorealistic style image as input), a good semantic correspondence between two images is needed.
    
    An interesting application of \cite{CNNMRF} can be found in \cite{SemanticStyleTransfer}, where a rough sketch can be transformed into a painting using high-level annotations of an example image. The sketch can have a very different layout of image parts like sky, water, forest, etc. with respect to the original image, and still be repainted into a very credible result.
    
    Used in unguided setting (without content image), the algorithm can produce high-quality textures based on a given example \cite{TextureSynthesis}. Moreover, using the style space from \cite{Gatys15} one can define an appropriate loss function and train a network that learns a given style \cite{TextureNetworks}. It alleviates the computational cost of generating a new texture (delegating it to the training stage). Basically, in \cite{TextureNetworks} the network learns to process input noise image in a way to minimize the style loss function introduced by \cite{Gatys15} for a chosen style image.
    
    In the following, we compare our approach with the original one \cite{Gatys15} and with the CNN-MRF method from \cite{CNNMRF}. The latter pursues similar goals to ours, improving the quality of style transfer via patch matching, while we suggest to enforce a tighter statistical constraint in an augmented style space.

\section{Helpful Modifications}\label{cm}
	Below we describe all modifications to the original style transfer algorithm in \cite{Gatys15} that turned out to be helpful. 
	
	Our core contributions are:
	\begin{enumerate}
		\item A better per-layer content/style weighting scheme (section \ref{weighting});
		\item Using more layers to capture more style properties (section \ref{layers});
		\item Using shifted activations when computing Gram matrices to eliminate sparsity and make individual entries more informative (section \ref{shift}) and also speed-up style transfer convergence;
		\item Targeting correlations of features belonging to different layers to capture more feature interactions (sections \ref{cor}, \ref{chain} and \ref{blur}).
	\end{enumerate}
	
	Modifications leading to questionable results are presented in section \ref{other}.
	
	The impact of our changes (both helpful and not) is presented in section  \ref{exp}.

	\subsection{Layer Weight Adjustment}\label{weighting}
		In \cite{Gatys15} content layer \texttt{conv4\_2} and 5 style layers $\left\{\texttt{conv1\_1}, \texttt{conv2\_1}, \texttt{conv3\_1}, \texttt{conv4\_1}, \texttt{conv5\_1}\right\}$ (of VGG-19) with unit weight $w_l$ each are used. We try to soften the ``content -- style'' separation by using the same set of layers for both content and style, but with a geometric weighting scheme: for each layer $l$ we set the style $w_l^s$ and content $w_l^c$ weights to
		
		$$w_l^s = 2^{D - d(l)},\,\,\,\, w_l^c = 2^{d(l)},$$
		
		where $D$ is the total number of layers used and $d(l)$ is the depth of layer $l$ with respect to all the other layers used. Thus we try to preserve more information about style and content (as neither is limited to being contained in a single layer) and also indicate that the most important style properties (e.g. colors, simple patterns) are captured by the bottom layers while content is mostly  represented by activations in the upper layers.
		
		For the setup described in section \ref{exp}, we have found this approach to sometimes produce a slight improvement in quality (in particular, the object tends to be better separated from the background), although oftentimes the results seem barely different.

	\subsection{Using More Layers}\label{layers}
		In \cite{Gatys15} only five layers are used for the calculation of Gram matrices. We try to enrich the style representation by calculating Gram matrices for all 16 convolutional layers of VGG-19. We observe that this approach leads to a consistent improvement in quality across the majority of style images considered.

	\subsection{Activation Shift}\label{shift}
		Convolutional layers of the normalized version \cite{VGGnorm} of VGG-19 output non-negative activations with mean $1$. For a typical image  outputs are also sparse: in all layers, each filter has usually few non-zero activations across spatial dimensions. This results in Gram matrices being sparse as well.
	
		We argue that sparsity is detrimental to style transfer quality. Precisely, it allows for unexpected patterns to appear in regions one would typically expect to be filled with, for example, a uniform background color.
	
		Our informal reasoning is as follows: zero entries of a Gram matrix are easy to misinterpret. Indeed, $\mathcal{G}^l_{i j} = 0$ could mean that features $i$ and $j$ of layer $l$ are either both absent, have one of them absent or both present but never appear together. And since, empirically, Gram matrices contain a lot of zero entries, they leave too much freedom for the optimization procedure to ``interpret'' them in a wrong way.
	
		We therefore decide to eliminate sparsity by calculating Gram matrices using  shifted activations: instead of setting
		
		$$\mathcal{G}^l = F^l{F^l}^T,$$
		
		(here $F^l$ is considered linearized into a 2D matrix of size $K_l \times X_l Y_l$) we put
		
		$$\mathcal{G}^l = \left(F^l + s\right) \left(F^l + s\right)^T,$$  
	
		where $s$ is the shift value added to matrices  element-wise. In this case, the gradient contributions in \cite{Gatys15} should be changed respectively:
		
		$$ \frac{\partial \mathcal{G}^l}{\partial F^l} = 2(F^l + s).$$
		
		Through several experiments we have concluded that putting $s = -1$ (i.e. centering activations at $0$) yields the best result. We have observed that shifting activations by this amount consistently improves the quality of style transfer across most of style images and style transfer methods. It also helps the optimization procedure to escape the starting point, where it can sometimes get stuck for hundreds of iterations otherwise.
		
		Finally, activation shift appears to promote uniform image repainting. When descending from content, the original style transfer often applies style with varying intensity, i.e. strongly repaints certain regions while barely modifying others (see, for example, second column of table \ref{UsVsThem2}). It may require fine-tuning the number of iterations and style weight to properly color all the regions without distorting content excessively. We encounter such problems less frequently when using an activation shift.

	\subsection{Correlations of Features from Different Layers}\label{cor}
		In \cite{Gatys15} style information is captured by a set of Gram matrices $\left\{\mathcal{G}^l\right\}$ of feature correlations within the same layer $l$. In our implementation we extend this definition to a set of Gram matrices $\left\{\mathcal{G}^{l k}\right\}$ of feature correlations belonging to possibly different layers $l$ and $k$. Since feature maps $F^l$ and $F^k$ may have different spatial dimensions, we always upsample the smaller map to match the size of the bigger one and thus define
		
		$$\mathcal{G}^{l k} = F^l \left[\textrm{up}(F^k)\right]^T,$$
		
		if $X_k \leqslant X_l$. $\mathcal{G}^{l k}$ has the size of $K_l \times K_k$ and, as earlier, an entry $\mathcal{G}^{l k}_{i j}$ represents the rate of co-occurrence of features $i$ and $j$ but now belonging to possibly different layers $l$ and $k$ respectively.
		
		Having this definition the derivatives with respect to activation volumes change if $l \neq k:$
		
		$$ \frac{\partial \mathcal{G}^{l k}}{\partial F^l} = \textrm{up}(F^k),\,\,\,\,\,\, \frac{\partial \mathcal{G}^{l k}}{\partial F^k} = \left[\textrm{up}^{-1}(F^l)\right]^T.$$
		
		With 16 convolutional layers in VGG-19 we have $2^{16^2}$ ways to ``tie'' them into a definition of style (i.e. a set of matrices $\mathcal{G}^{l k}$). One of the approaches we tried first was to tie all layers with a single content layer, motivated by our perception of style as of the way in which low-level features correlate with high-level features. Precisely, instead of using $\left\{\mathcal{G}^l\,|\, l = 1 \dots 16\right\}$ (with layers \texttt{conv1\_1} to \texttt{conv5\_4} numbered from 1 to 16) as our style representation, we use $\left\{\mathcal{G}^{c l}\,|\, l = 1 \dots 16\right\}$ with $c$ corresponding to a single high-level layer (\texttt{conv4\_2} for example). However, as seen in section \ref{exp}, this particular style representation didn't yield any improvement. In general, we observe that tying distant in terms of depth layers produces poor results.

	\subsection{Correlation Chain}\label{chain}
		In the framework of using inter-layer correlations for style as described in section \ref{cor}, we now consider the following ``chained'' style representation: $\left\{\mathcal{G}^{l, l-1}\,|\,l = 2 \dots 16\right\}$. We thus still constrain correlations of high- and low-level features, but in a local way, where only the correlation with immediate neighbors are considered.
		
		This approach has lead to a consistent and often significant improvement in style transfer quality in most cases considered.

	\subsection{Blurred Correlations}\label{blur}
		Recall that in section \ref{cor} the smaller activation map is upsampled to match the bigger one and then pixel-wise correlations of their features are calculated. However, even having the same spatial dimensions, feature maps still correspond to features of different scales. For example, consider layers \texttt{conv1\_1} and \texttt{conv1\_2}: they have the same size, but features of \texttt{conv1\_2} are composed of $3\times 3$ patches of \texttt{conv1\_1}-features (as VGG-19 uses $3\times 3$ kernels). Therefore, it may be reasonable to not capture the correlation of \texttt{conv1\_2}-activations the activations in pixels directly ``underneath'' them (which is the case in the default implementation of $\mathcal{G}^{l k}$), but rather measure its correlation with an average activation of the other feature in the respective $3\times 3$-patch.
		
		Precisely, we implement this as follows:
		$$\mathcal{G}^{l k} = F^l \left[\textrm{blur}^{l - k} \circ \textrm{up}(F^k)\right]^T,$$
		i.e. we apply box blurring  $l - k$ times to the upsampled version of the smaller activation. Then the new derivatives become
		
		$$ \frac{\partial \mathcal{G}^{l k}}{\partial F^l} = \textrm{blur}^{l - k} \circ \textrm{up}(F^k),\,\,\,\,\,\, \frac{\partial \mathcal{G}^{l k}}{\partial F^k} = \left[\textrm{up}^{-1}\circ \textrm{blur}^{l - k}\left(F^l\right)\right]^T.$$
		
		This modification has produced mixed, but overall positive results. However, it does complicate the objective function which results in slow and unreliable convergence.

\section{Other Approaches}\label{other}
	Below we describe some other attempts to improve style transfer that didn't work out in the end.
	
	\subsection{Gradient Masking}\label{gradmask}
		One of our first attempts to improve style transfer quality was to prevent unexpected textures from appearing in the background of a content image.
		
		We try a very simple idea: apply a binary mask to the style gradient from every layer. The mask was supposed to suppress undesirable textures in areas with weak content response, corresponding to monotone areas. Precisely, we replace the derivative:
		
		$$ \frac{\partial \mathcal{G}^l}{\partial F^l} = 2 F^l$$
		
		with
		
		$$ \frac{\partial \mathcal{G}^l}{\partial F^l} = 2 F^l\odot S^l,$$
		
		with mask $S$ being generated by simple thresholding of the feature maps of the content image $P^l$:
		
		$$S^l_{i}(x, y) = \begin{cases}
		1 & \textrm{ if } P^l_{i} (x, y) >  \texttt{threshhold}(l)\\
		0 & \textrm{ otherwise}
		\end{cases}.$$
		
\begin{figure}
			\centering
			\includegraphics[resolution=300]{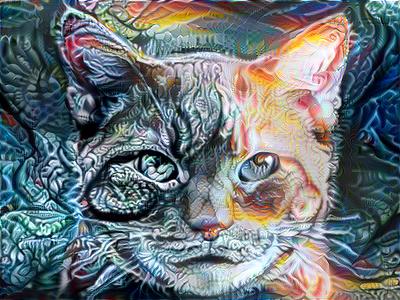}
			\includegraphics[resolution=300]{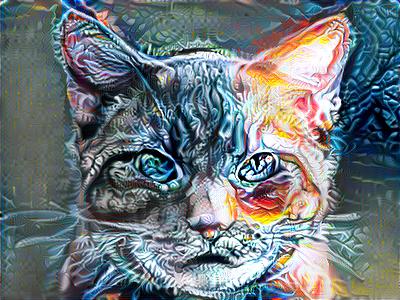}\\
			\includegraphics[resolution=300]{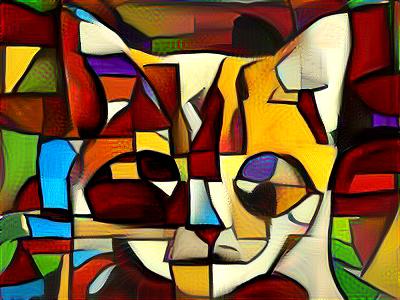}
			\includegraphics[resolution=300]{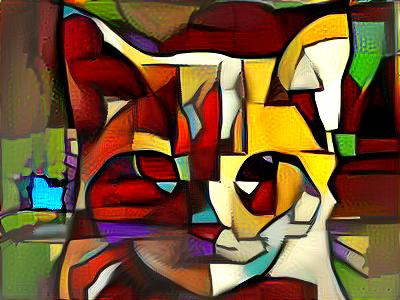}\\
			\includegraphics[resolution=300]{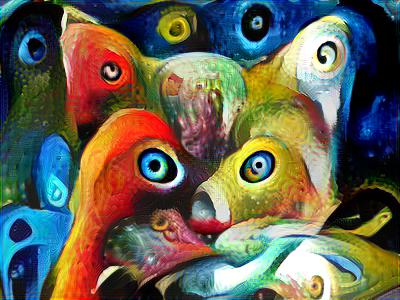}
			\includegraphics[resolution=300]{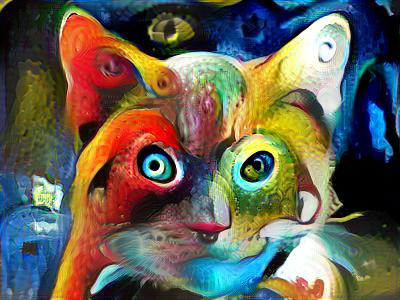}\\
			\includegraphics[resolution=300]{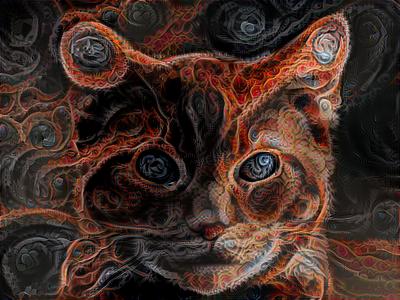}
			\includegraphics[resolution=300]{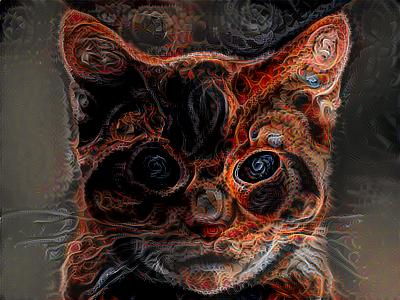}\\
			\includegraphics[resolution=300]{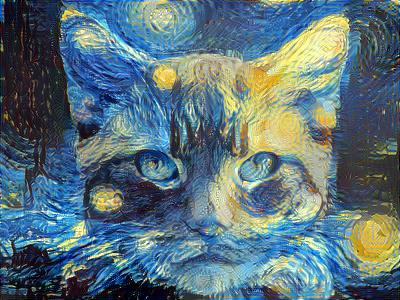}
			\includegraphics[resolution=300]{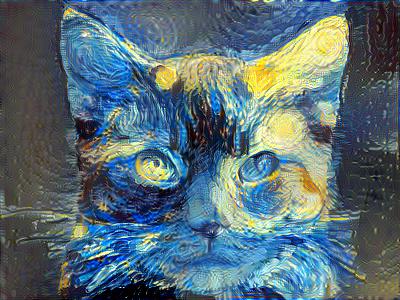}\\
			\includegraphics[resolution=300]{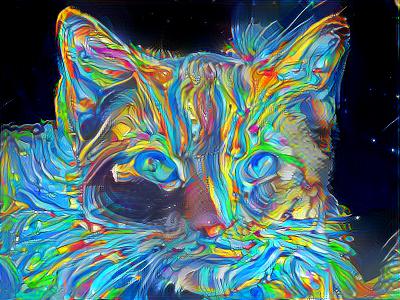}
			\includegraphics[resolution=300]{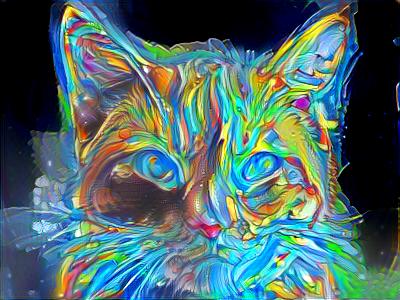}\\
			\includegraphics[resolution=300]{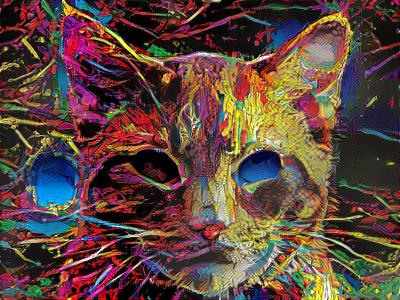}
			\includegraphics[resolution=300]{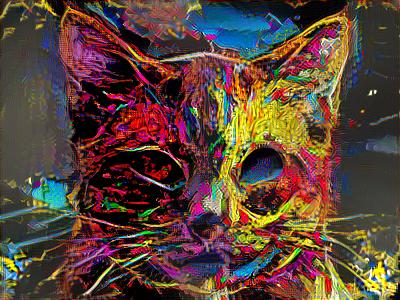}
			\caption{Comparison of the original algorithm in \cite{Gatys15} (left) and the algorithm with masked style gradient (right). Content image (here used in $400\times 300$ resolution): \cite{cat}. Style images (top to bottom): \cite{face, cubism, eyes, fractals, gogh, cat3, skull}.}
			\label{masked}
		\end{figure}

		The \texttt{threshhold}$(l)$ value was set in a way to keep the desired portion of the output. From a few experiments we have found that keeping $100\%, 40\%, 20\%, 10\%, 10\%$ of biggest entries for the layers \texttt{conv1\_1, conv2\_1, conv3\_1, conv4\_1, conv5\_1} respectively indeed suppressed texture generation on the background while preserving reasonable quality of style transfer to the object in the foreground. Notice that we let the mask pass more shallow layer activations and fewer deep layer activations, expecting the color and simple style features to replace the background in the content image. Examples of style transfer results using masked gradients are presented in figure \ref{masked}. 
		
		However, applying a mask modifies the gradient and thus damages the minimization procedure. In practice we observed that after several hundreds of iterations the error began to stall or even increase. Also, the portion of elements one should set to zero was found experimentally for a given content image, and we do not expect this approach to  generalize well.

	\subsection{Amplifying Activations}\label{ampl}
		In an attempt to make the approach in section \ref{gradmask} generalize better, we decide to not have a hard threshold on activations, but rather amplify them so that regions with strong feature responses in the content image are prioritized. Precisely, we set
		
		$$\mathcal{G}^l = {F^l}^p \left({F^l}^p\right)^T,$$
	
		so that the derivative becomes
		
		$$\frac{\partial \mathcal{G}^l}{\partial F^l} = 2 p {F^l}^{2 p - 1},$$
		
		where the exponentiation is performed element-wise. Then, setting $p$ to $2$ would result in $\frac{\partial \mathcal{G}^l}{\partial F^l} = 2 p {F^l}^{3},$ amplifying and prioritizing change in strong activation regions.
		
		With this objective we indeed observed how during style transfer feature-rich regions (e.g. eyes) were modified before and more strongly than the rest of the image. However, we could not fully evaluate this idea (and, notably, see if this modified objective makes sense) as it had serious problems with convergence. We think the reason is the objective function becoming highly non-linear, since convergence problems were less severe as we reduced $p$ from $2$ down to $1$.
	
	\subsection{Adjacent Activations Correlations}\label{adj}
		In \cite{Gatys15}, as well as in previous sections, pixel-wise feature correlations are considered (apart from a slight modification in section \ref{blur}, where high-level features are correlated with average values of low-level feature patches). We now try to also constrain how features correlate with other features not only in the exact same spatial location, but in the 8 adjacent locations as well. Precisely, a $K_l \times K_k$ Gram matrix $\mathcal{G}^{l k}$ expands into a $3\times 3\times K_l \times K_k$ matrix with new dimensions corresponding to shifts along $X$ and $Y$ axis ($-1, 0, 1$). If we introduce the following matrix of pixel shifts
		
		$$S = \left[\begin{matrix}
		(-1, -1) & (-1, 0) & (-1, 1) \\
		(0, -1) & (0, 0) & (0, 1) \\
		(1, -1) & (1, 0) & (1, 1) \\
		\end{matrix}\right],$$
		
		then we define the new Gram matrix as
		
		$$\tilde{\mathcal{G}}^{l k} = \textrm{map}\left(\mathcal{G}^{l k}, S\right) = \left[\begin{matrix}
		\mathcal{G}^{l k}(-1, -1) & \mathcal{G}^{l k}(-1, 0) & \mathcal{G}^{l k}(-1, 1) \\
		\mathcal{G}^{l k}(0, -1) & \mathcal{G}^{l k}(0, 0) & \mathcal{G}^{l k}(0, 1) \\
		\mathcal{G}^{l k}(1, -1) & \mathcal{G}^{l k}(1, 0) & \mathcal{G}^{l k}(1, 1) \\
		\end{matrix}\right],$$
		
		where
		$$\mathcal{G}^{l k} (dx, dy) = F^l \textrm{shift}(F^k, dx, dy)^T,$$
		
		with the ``shift'' operation shifting (with 0 padding) the activation map $dx$ and $dy$ pixels along $X$ and $Y$ dimensions.
		
		With this new construction the derivatives with respect to activation maps become
		
		$$ \frac{\partial \tilde{\mathcal{G}}^{l k}}{\partial F^l} = \textrm{map}\left(\textrm{shift}({F^k}^T,\cdot,\cdot), S\right),\,\,\,\,\,\,\frac{\partial \tilde{\mathcal{G}}^{l k}}{\partial F^k} = \textrm{map}\left(\textrm{shift}(F^l, \cdot, \cdot), -S\right).$$
		
		This modification has a significant impact on the style transfer result, but in a negative way more often than not.

	\subsection{Content-aware Gram Matrices}\label{cubes}
		We now describe an attempt to make the style more ``content-aware''. Precisely, during style transfer, we want different styles to be applied to different regions of an image, depending on the content of the region. In a simple case we want styles of background and foreground to be translated separately (contrary to being blended into a single statistic over the whole image). In a more complex case we want the style to vary as a function of content and result in an overall improvement of style transfer quality.
		
		We first pick a content layer, for example $l_c = \texttt{conv4\_2}$. It has $K_{l_c} = 512$ activation maps $F^{l_c}_k$, i.e. 512 ``types of content'' that we distinguish. For each type we want to have separate styles, i.e. separate sets of Gram matrices $\mathcal{G}^{l}$ (for whichever $l$s we decide to represent style).
		
		We store this information by introducing depth to Gram matrices: a third dimension of size $K_{l_c}$. We thus introduce ``content-aware Gram matrices'' $\mathcal{G}^{l}_{l_c}$ of size $K_{l_c} \times K_l \times K_l$ as
		
		$$\left(\mathcal{G}^{l}_{l_c}\right)_{k,\cdot,\cdot} = \texttt{styleBy}\left(F^l, F^{l_c}_k\right),$$
		
		where \texttt{styleBy} is a function used to compute a style statistic of $F_l$ that best describe content of type $k$ based on the supplied activation map $F^{l_c}_k$. Having non-negative activations (see section \ref{shift}) we suggest defining \texttt{styleBy} as
		
		$$\left[\texttt{styleBy}\left(F^l, F^{l_c}_k\right)\right]_{i j} = \sum_{x, y} F^l_i (x, y) \cdot F^l_j (x, y) \cdot F^{l_c}_k (x, y),$$
		
		i.e. the $K_{l_c}$ slices of $\mathcal{G}^{l}_{l_c}$ are Gram matrices with correlation contributions weighted by response of feature $k$ in layer $l_c$ (for simplicity it is assumed here that $F^l$ and $F^{l_c}$ have the same spatial dimensions, which is generally not true. In this case the smaller map is upsampled as described in section \ref{cor}).
		
		Thus each feature collects style statistics present in regions where the feature has a high response.
		
		Formally
		
		$$\left(\mathcal{G}^{l}_{l_c}\right)_{k,i,j} = \left(F^l_i \odot {F^{l_c}_k}\right){F^l_j}^T,$$
		
		and the derivatives are
		
		$$ \frac{\partial \left(\mathcal{G}^{l}_{l_c}\right)_{k,i,j}}{\partial F^l_i} = F^l_j \odot {F^{l_c}_k},\,\,\,\,\,\,\frac{\partial \left(\mathcal{G}^{l}_{l_c}\right)_{k,i,j}}{\partial F^l_j} = F^l_i \odot {F^{l_c}_k},\,\,\,\,\,\,\frac{\partial\left(\mathcal{G}^{l}_{l_c}\right)_{k,\cdot,\cdot}}{\partial F^{l_c}_k} = F^l_i \odot F^l_j.$$
		
		On a practical note, we have also observed that shifting style layer activations (see section \ref{shift}) was helpful in this context as well.
		
		We were only able to test this approach on $256\times 256$ images with \texttt{conv4\_2} as the content layer and \texttt{conv1\_1}, \texttt{conv2\_1},\texttt{conv3\_1}, \texttt{conv4\_1} (almost as in \cite{Gatys15}) as style layers. While in the photorealistic context it has lead to an improvement (see table \ref{cubesTable}), it did not generalize well to other styles and generally performed comparably to other statistic-based approaches (while being much slower).
	
		\begin{table}[]
			\centering
			\hspace*{-2cm}\begin{tabular}{ccccc}
				Style & Gatys et al. \cite{Gatys15} & Li and Wand \cite{CNNMRF} & Chain Blurred & Content-aware \\

				\includegraphics[width=3cm]{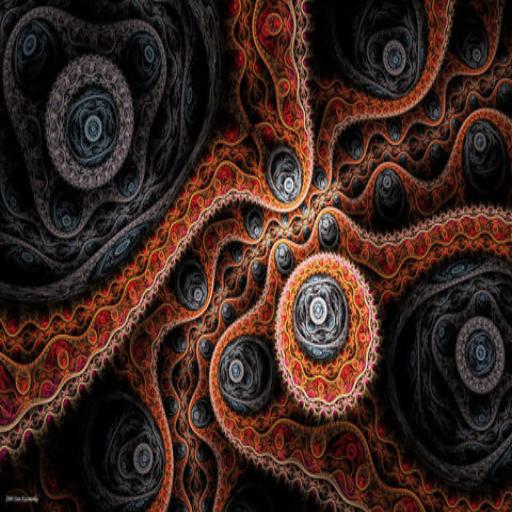} & \includegraphics[width=3cm]{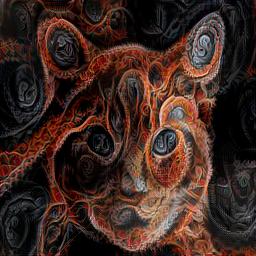} & \includegraphics[width=3cm]{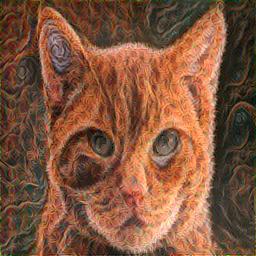} & \includegraphics[width=3cm]{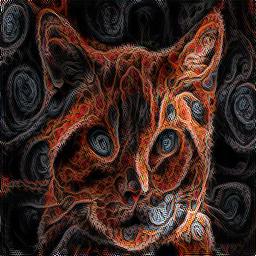} &
				\includegraphics[width=3cm]{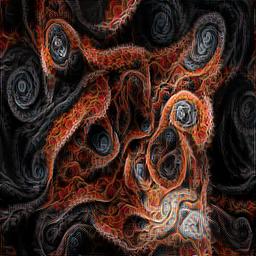} \\
				
				\includegraphics[width=3cm]{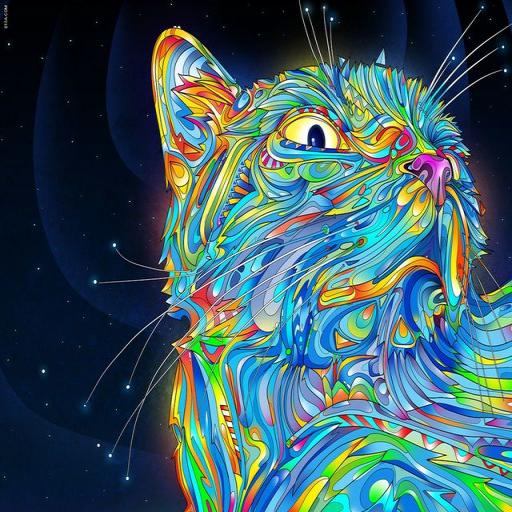} & \includegraphics[width=3cm]{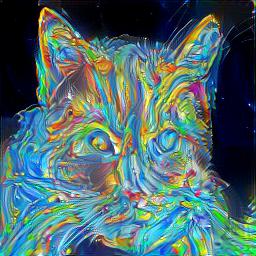} & \includegraphics[width=3cm]{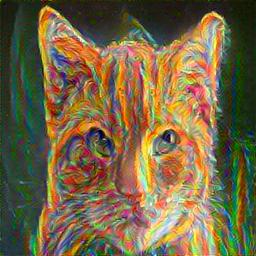} & \includegraphics[width=3cm]{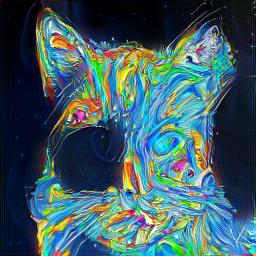} &
				\includegraphics[width=3cm]{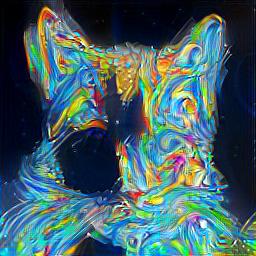} \\
				
				\includegraphics[width=3cm]{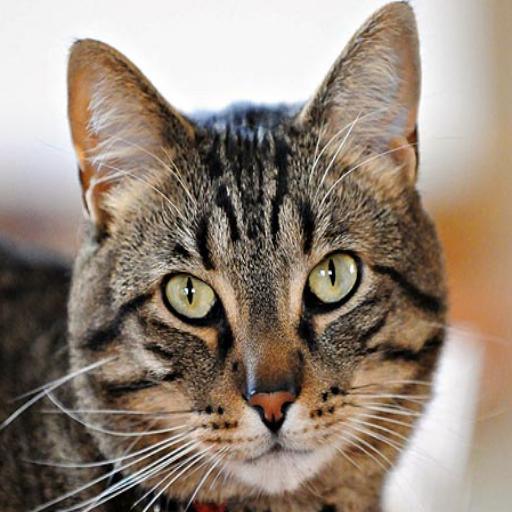} & \includegraphics[width=3cm]{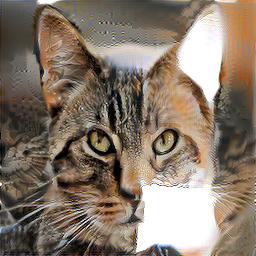} & \includegraphics[width=3cm]{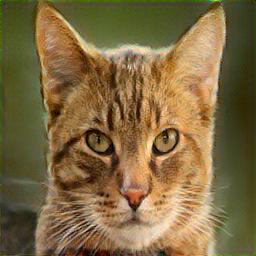} & \includegraphics[width=3cm]{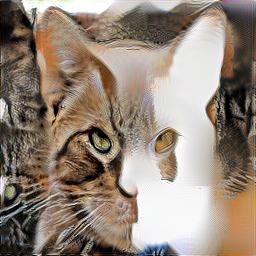} &
				\includegraphics[width=3cm]{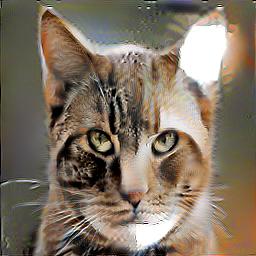}
			\end{tabular}
			\caption{Content-aware Gram matrices mixed performance compared to other approaches on $256\times 256$ images. Content: figure \ref{content}. Style images (top to bottom): \cite{fractals, cat3}.}
			\label{cubesTable}
		\end{table}

	\subsection{Gram Cubes}
		In the spirit of section \ref{cubes} we can also try to define style as a set of ``Gram cubes'' $\tilde{\mathcal{G}}^l$ of size $K_l \times K_l \times K_l$ storing triple correlations of features in each layer:
		
		$$\left(\tilde{\mathcal{G}}^{l}\right)_{k,i,j} = \left(F^l_i \odot F^l_j\right){F^l_k}^T.$$
		
		In practice, however, it did not lead to any promising results and was not deemed worth the computational complexity.

\section{Experiments}\label{exp}
	Suggestions described in section \ref{cm} were implemented in Torch \cite{Torch} on top of Kai Sheng Tai's implementation \cite{tai} of the original style transfer \cite{Gatys15}. The network used was VGG-19 \cite{VGG}, precisely, the normalized version with weights scaled to have a unit mean neuron activations over images and positions \cite{VGGnorm}.
	
	All style and content images were rescaled to the size of $512\times 512$ pixels. For each presented setup $270$ iterations of L-BFGS optimization algorithm were performed starting from the content image. 
	
	By default the style weight was set to $2\times 10^9$. However, in some cases the algorithm had trouble escaping the starting image for hundreds of iterations, in which case we either modified the style weight and re-run the computation or saved a result of a later iteration. The CNN-MRF \cite{CNNMRF} images were generated using default settings of the implementation shared by authors in \cite{CNNMRFcode}.
	
\begin{figure}[h]
		\centering
		\includegraphics[resolution=300]{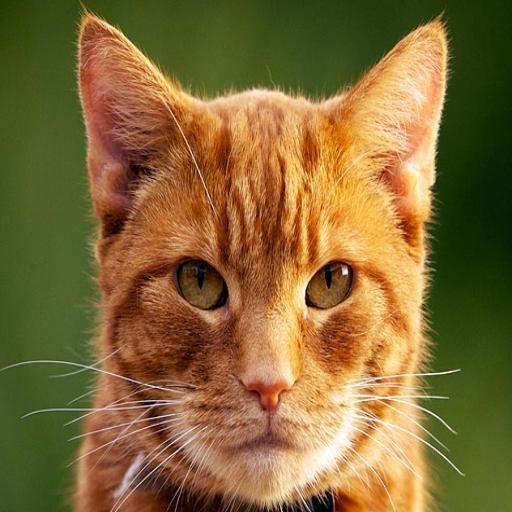}
		\caption{Content image \cite{cat} considered for experiments.}
		\label{content}
	\end{figure}

	The content image is displayed in figure \ref{content}.
	
	Style transfer methods compared in this work are presented in table \ref{methods}. In tables \ref{UsVsThem1} and \ref{UsVsThem2} we compare \texttt{Classic} \cite{Gatys15}, \texttt{CNN-MRF} \cite{CNNMRF} and \texttt{Chain Blurred} to demonstrate the performance of our approach compared to existing solutions. In table \ref{UsVsUs} we compare all configurations to demonstrate the impact of individual modifications described in section \ref{cm}.

	\begin{table}[]
		\centering
		\begin{tabular}{|p{1.3cm}|p{0.9cm}|p{0.9cm}|p{0.9cm}|p{1.5cm}|p{1cm}|p{1cm}|p{1cm}|p{1cm}|p{1cm}|p{1cm}|}
			\hline
			Method name                      & Classic                       & Classic Shifted                       & Classic Dense     & All-To-Content                                                                            & Chain         & Chain Uniform         & Chain Unshifted         & Chain Blurred        & Chain Extended        & CNN-MRF                           \\ \hline
			Content layers                   & \multicolumn{2}{p{2cm}|}{conv4\_2}                                         & \multicolumn{7}{p{4cm}|}{all convolutional}                                                                                                                                                                                         & \multirow{6}{*}{Default} \\ \cline{1-10}
			Style layers                     & \multicolumn{2}{p{2cm}|}{conv1\_1, conv2\_1, conv3\_1, conv4\_1, conv5\_1} & all convolutional & conv4\_2 - conv5\_4, conv4\_2 - conv5\_3, \dots, conv4\_2 - conv1\_2, conv4\_2 - conv1\_1 & \multicolumn{5}{p{6cm}|}{conv5\_4 - conv5\_3, conv5\_3 - conv5\_2, \dots, conv2\_1 - conv1\_2, conv1\_2 - conv1\_1} &                                   \\ \cline{1-10}
			Weighting scheme                 & \multicolumn{2}{l|}{uniform}                                          & \multicolumn{3}{p{2cm}|}{geometric}                                                                                                & uniform               & \multicolumn{3}{p{2cm}|}{geometric}                                         &                                   \\ \cline{1-10}
			Activation shift                 & 0                             & \multicolumn{5}{l|}{-1}                                                                                                                                                                       & 0                       & \multicolumn{2}{l|}{-1}                      &                                   \\ \cline{1-10}
			Blurred correlation              & \multicolumn{7}{l|}{no}                                                                                                                                                                                                                                 & yes                  & no                    &                                   \\ \cline{1-10}
			Adjacent activations correlation & \multicolumn{8}{l|}{no}                                                                                                                                                                                                                                                        & yes                   &                                   \\ \hline
		\end{tabular}
		\caption{Style transfer methods compared.}
		\label{methods}
	\end{table}
	
	\begin{table}[]
		\centering
		\hspace*{-2cm}\begin{tabular}{cccc}
			Style & Classic (Gatys et al. \cite{Gatys15}) & CNN-MRF (Li and Wand \cite{CNNMRF}) & Chain Blurred (ours) \\
			\includegraphics[width=4cm]{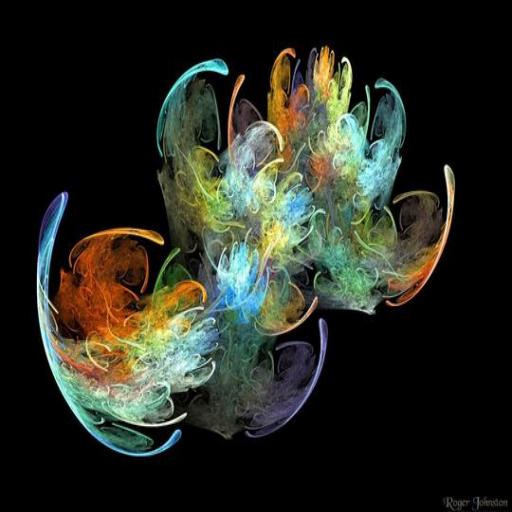}      &   \includegraphics[width=4cm]{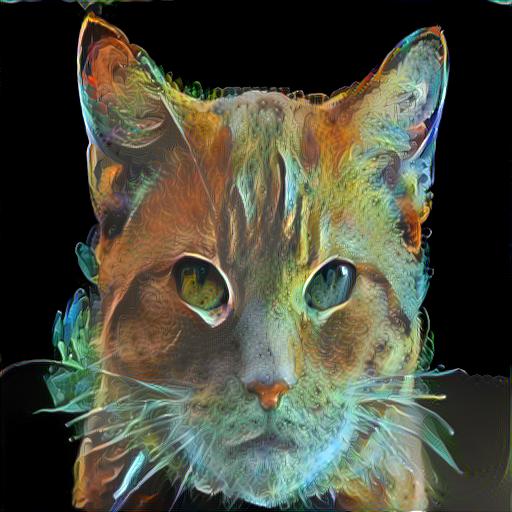}                     & \includegraphics[width=4cm]{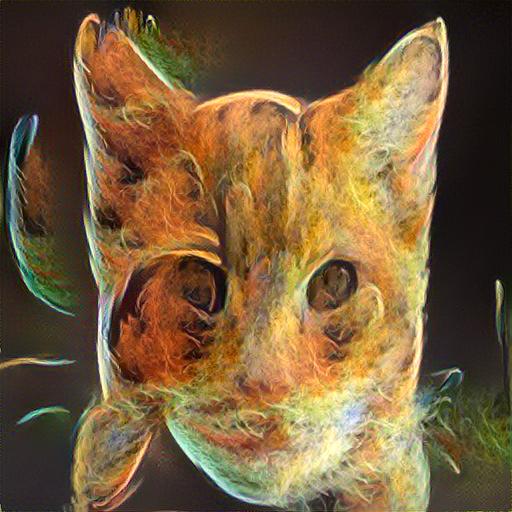}                      &  \includegraphics[width=4cm]{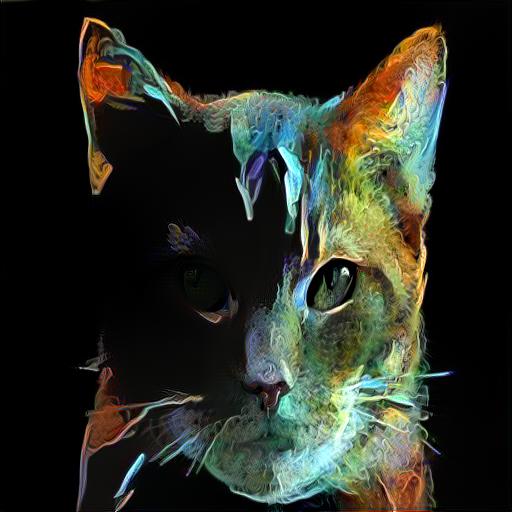}            \\
			
			\includegraphics[width=4cm]{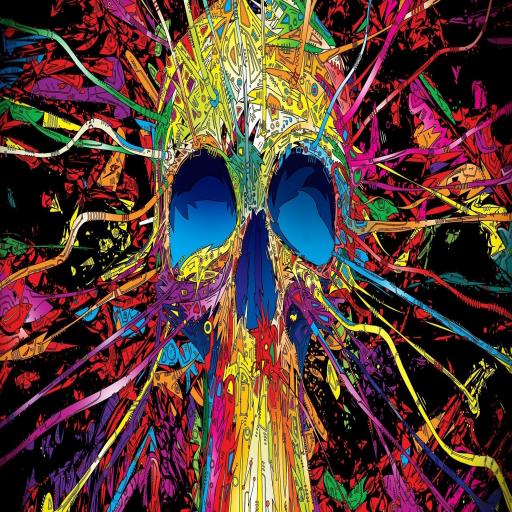}      &   \includegraphics[width=4cm]{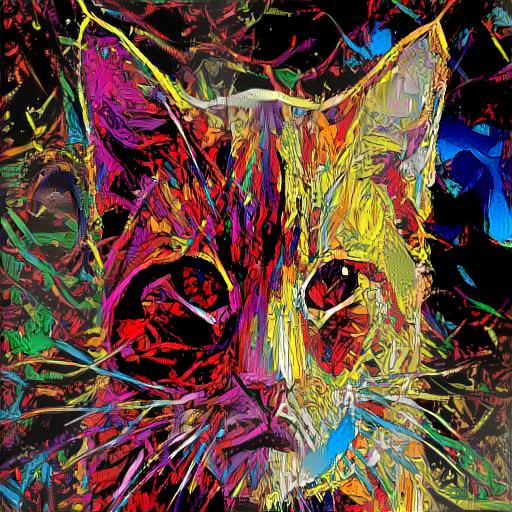}                     & \includegraphics[width=4cm]{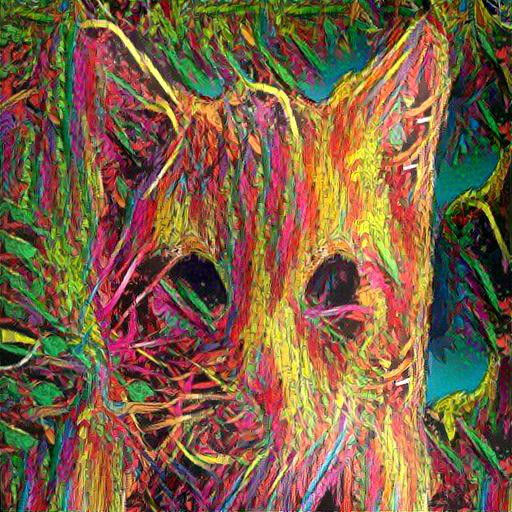}                      &  \includegraphics[width=4cm]{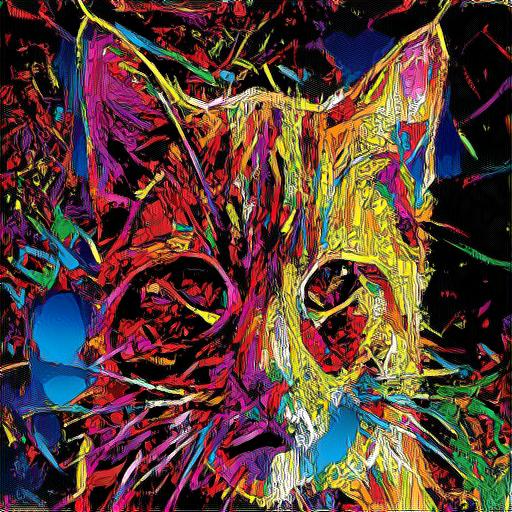}            \\
			
			\includegraphics[width=4cm]{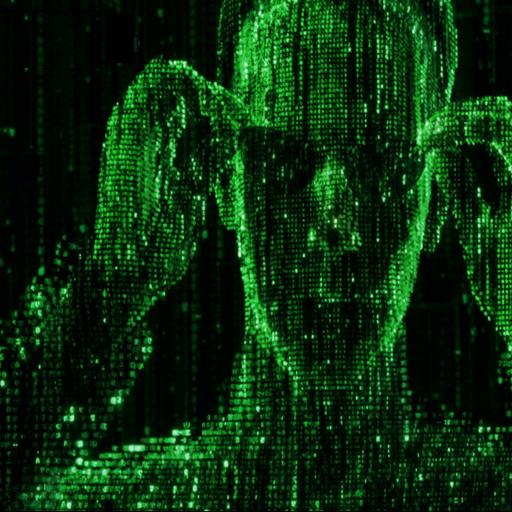}      &   \includegraphics[width=4cm]{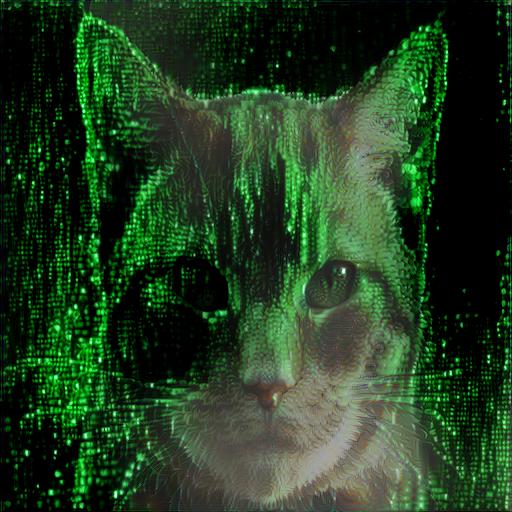}                     & \includegraphics[width=4cm]{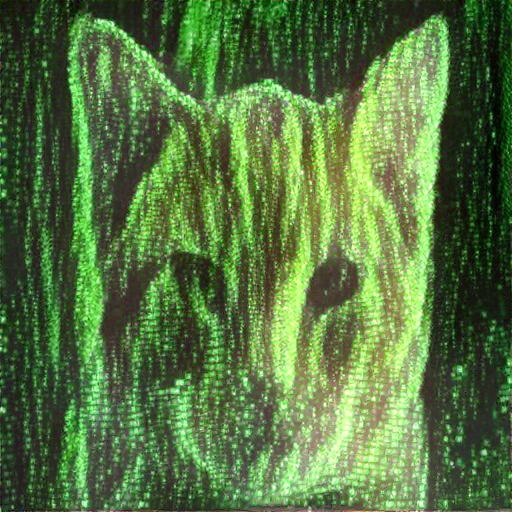}                      &  \includegraphics[width=4cm]{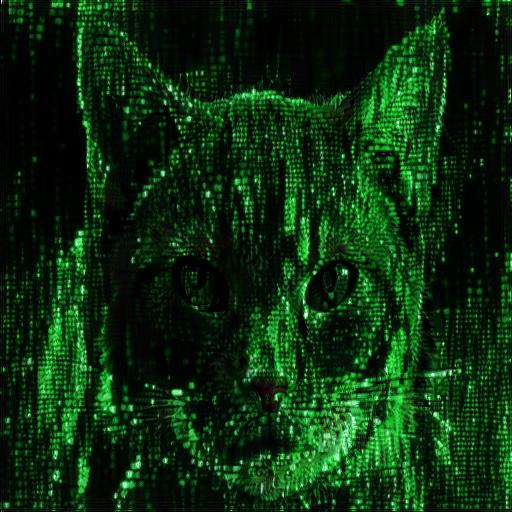}            \\
			
			\includegraphics[width=4cm]{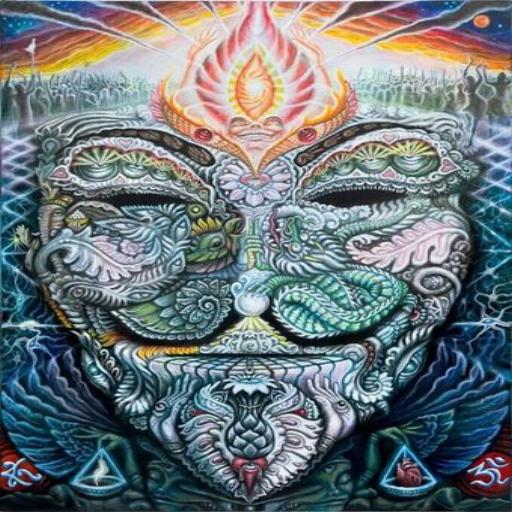}      &   \includegraphics[width=4cm]{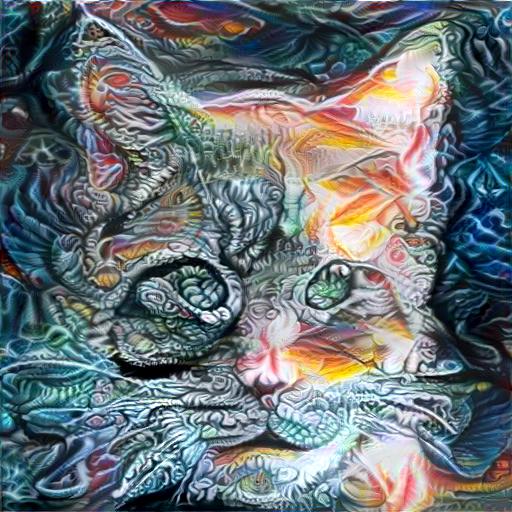}                     & \includegraphics[width=4cm]{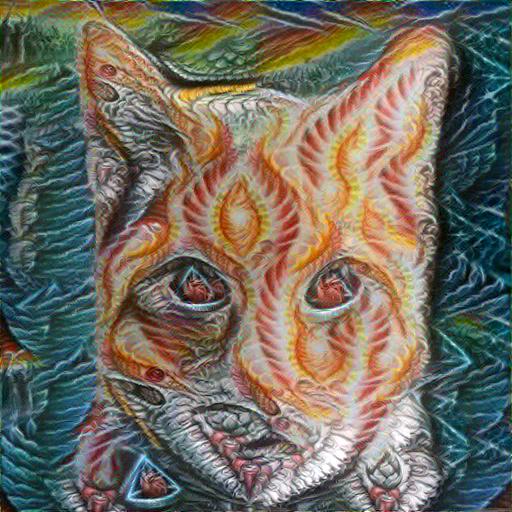}                      &  \includegraphics[width=4cm]{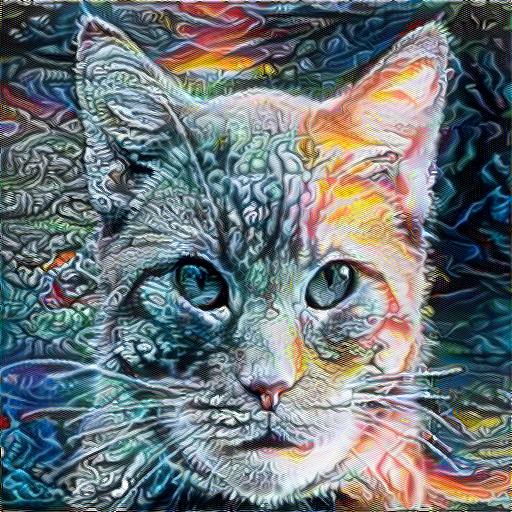}            
		\end{tabular}
		\caption{Comparing our results to existing style transfer solutions (part 1). Style images (top to bottom): \cite{f2, skull, matrix, face}.}
		\label{UsVsThem1}
	\end{table}

	\begin{table}[]
		\centering
		\hspace*{-2cm}\begin{tabular}{cccc}
			Style & Classic (Gatys et al. \cite{Gatys15}) & CNN-MRF (Li and Wand \cite{CNNMRF}) & Chain Blurred (ours) \\
			\includegraphics[width=4cm]{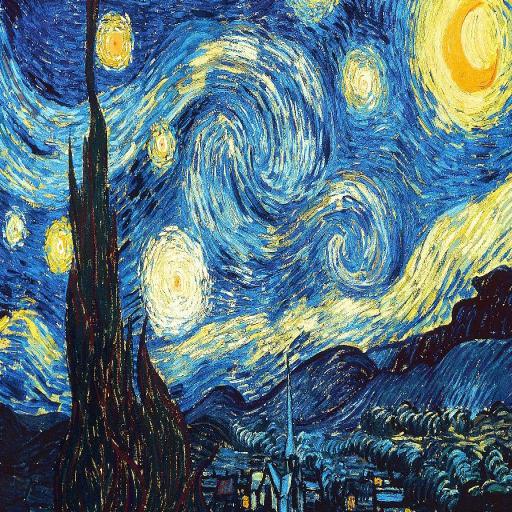}      &   \includegraphics[width=4cm]{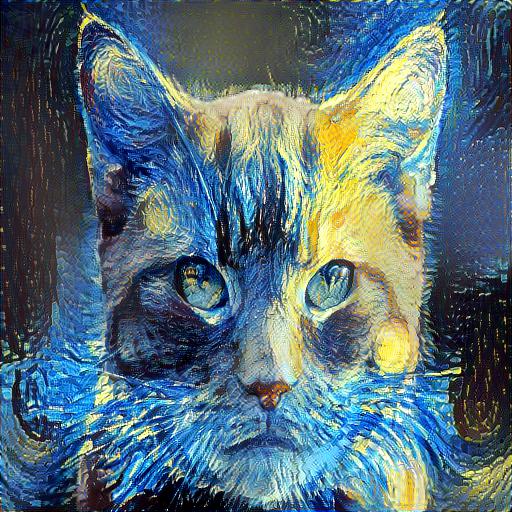}                     & \includegraphics[width=4cm]{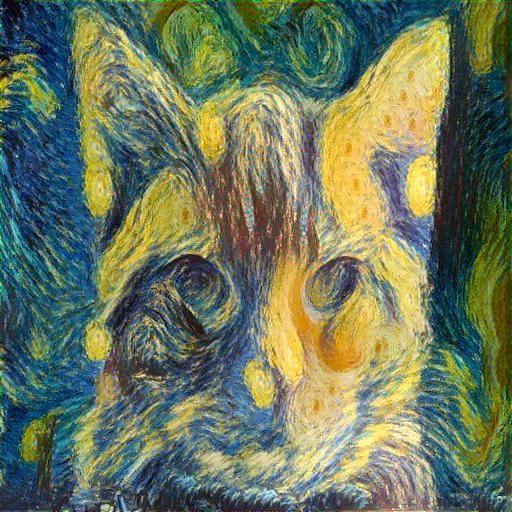}                      &  \includegraphics[width=4cm]{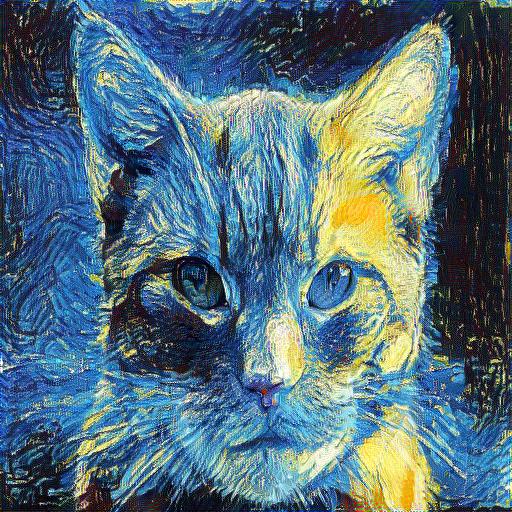} \\
			
			\includegraphics[width=4cm]{styles/fractals}      &   \includegraphics[width=4cm]{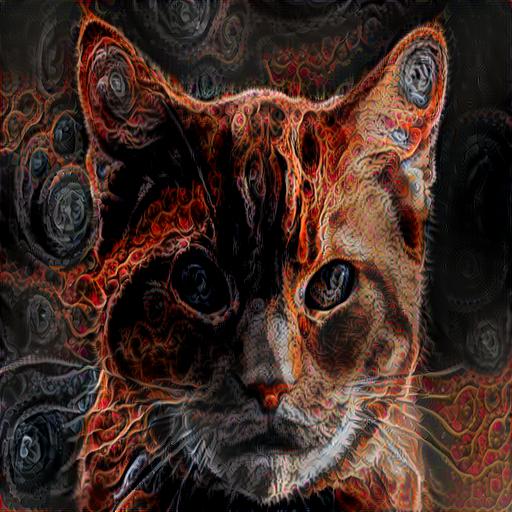}                     & \includegraphics[width=4cm]{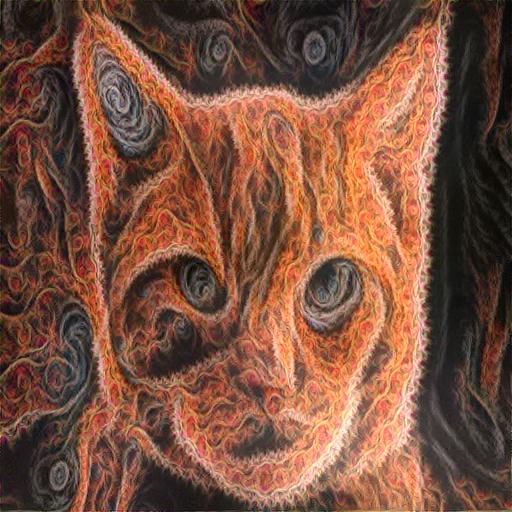}                      &  \includegraphics[width=4cm]{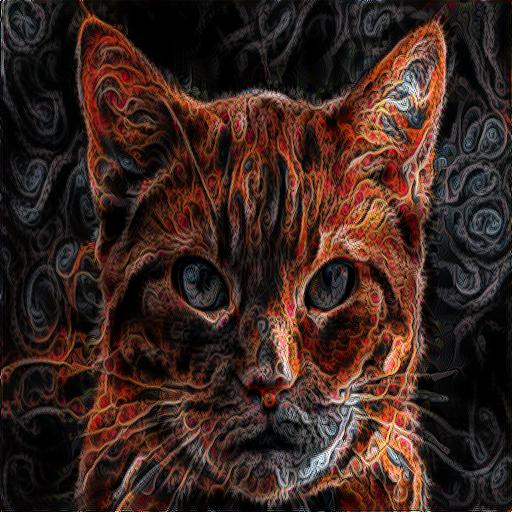} \\
			
			\includegraphics[width=4cm]{styles/cat3}      &   \includegraphics[width=4cm]{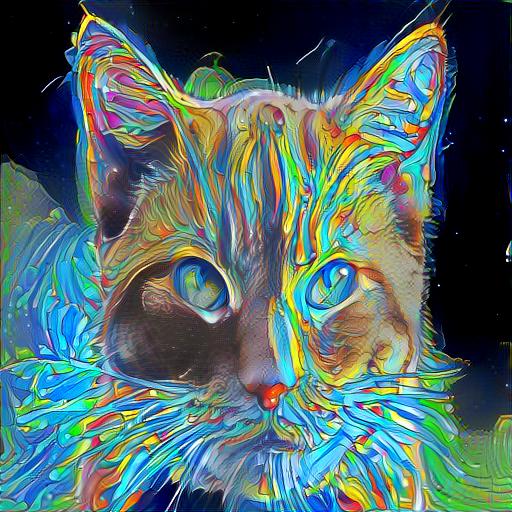}                     & \includegraphics[width=4cm]{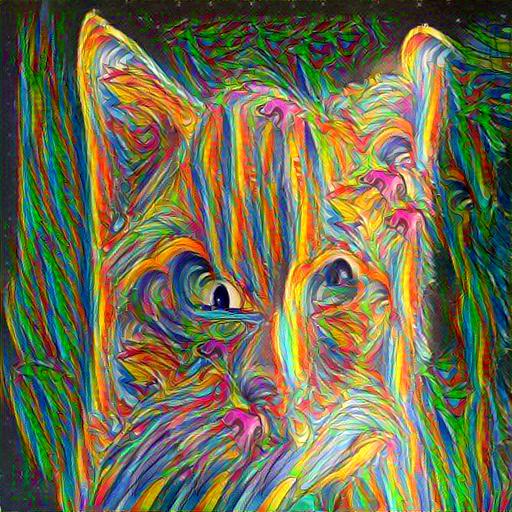}                      &  \includegraphics[width=4cm]{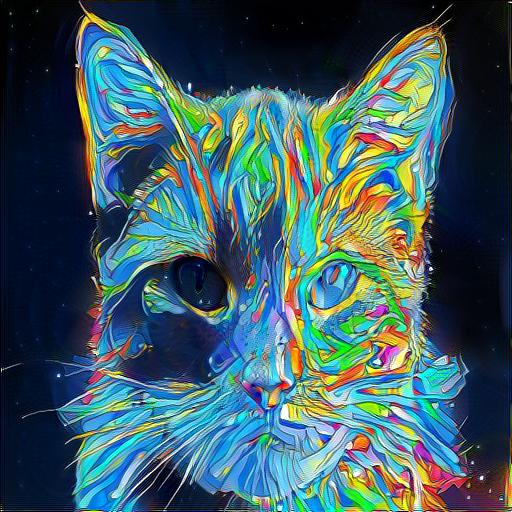} \\
			
			\includegraphics[width=4cm]{styles/cat2}      &   \includegraphics[width=4cm]{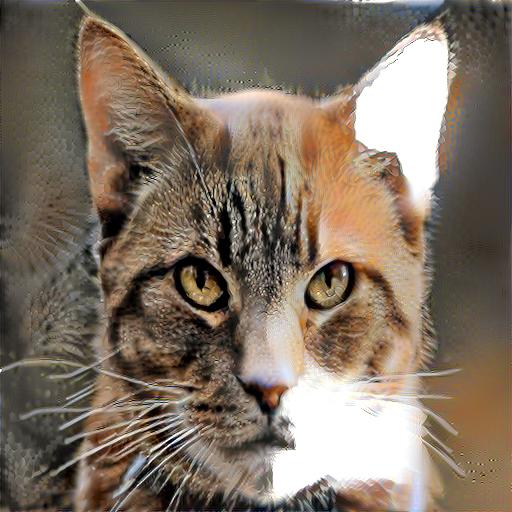}                     & \includegraphics[width=4cm]{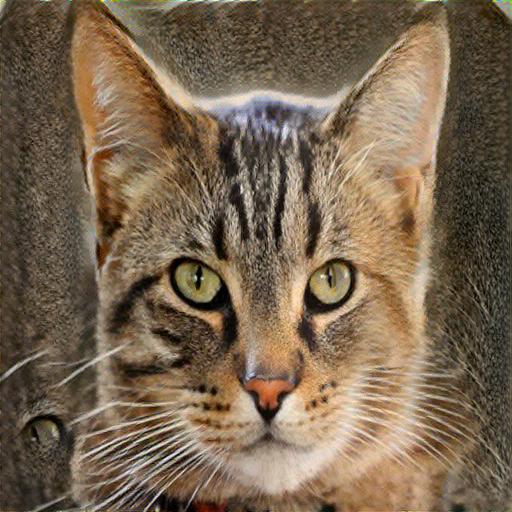}                      &  \includegraphics[width=4cm]{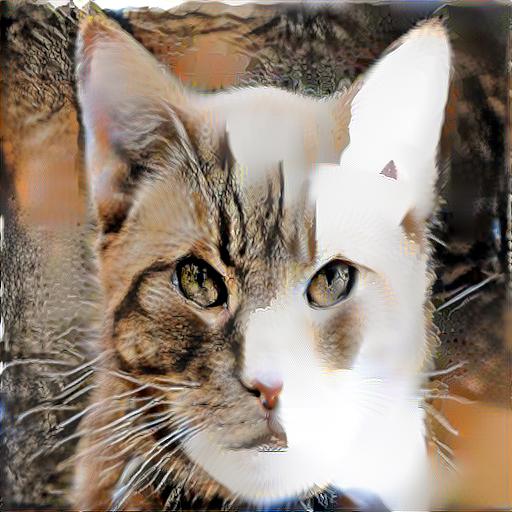} \\
		\end{tabular}
		\caption{Comparing our results to existing style transfer solutions (part 2). Style images (top to bottom): \cite{gogh, fractals, cat3, cat2}.}
		\label{UsVsThem2}
	\end{table}

\begin{landscape}
	\includepdf[ pagecommand={\thispagestyle{plain}}]{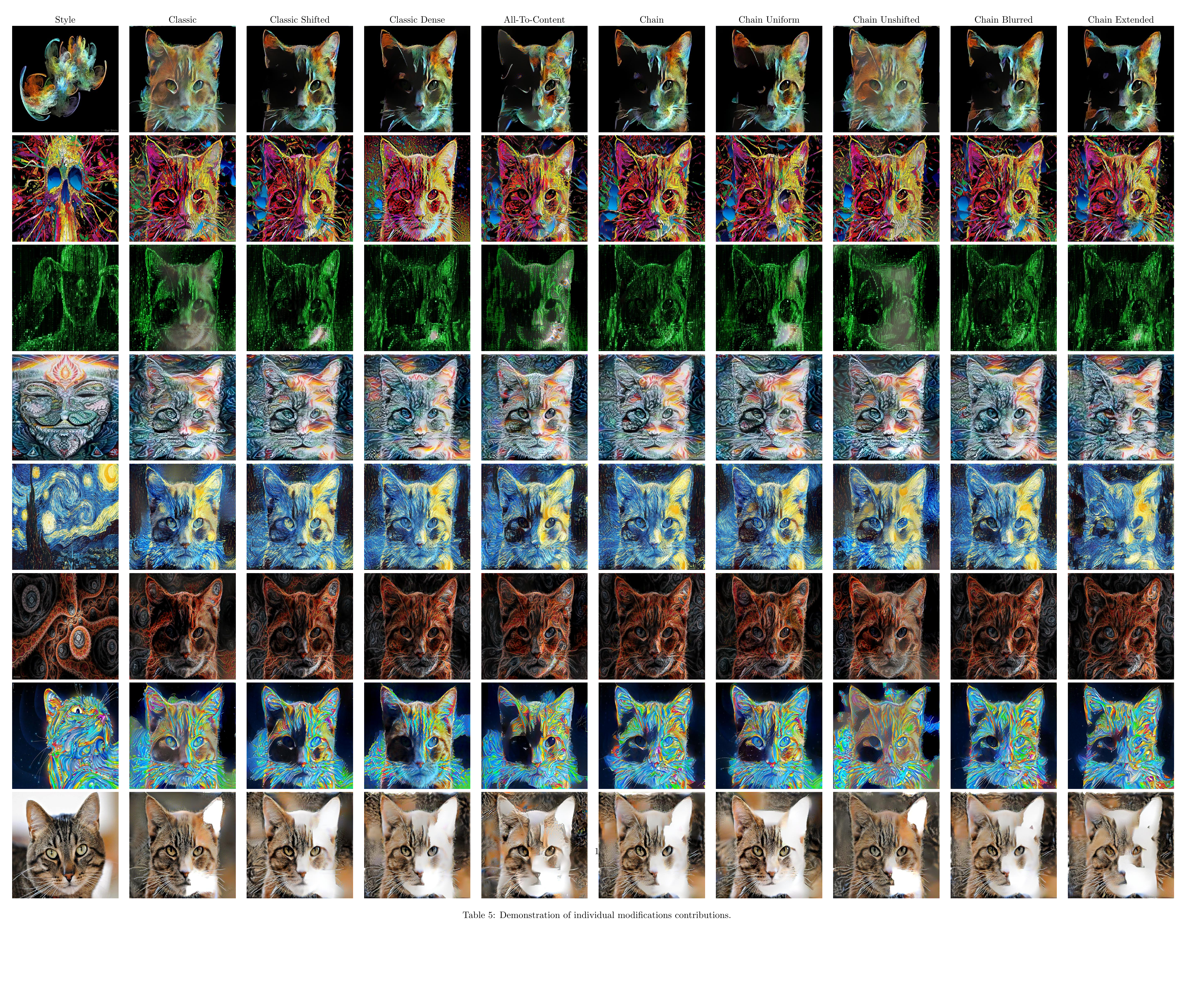}		\label{UsVsUs}
\end{landscape}

\section{Conclusion}\label{conc}
	In this work several ways to improve the style transfer algorithm suggested in \cite{Gatys15} were investigated. The direction of our experiments was mostly inspired by the Julesz Conjecture \cite{Julesz, Simoncelli}, according to which two textures sharing a certain set of statistics should be visually indistinguishable. Similarly, our conjecture was that there exists a set of statistics describing the style of an image with desirable visual properties, and we used feature correlation matrices from \cite{Gatys15} as a starting point.

	Our most useful contributions turned out to be, in  order of decreasing impact
	\begin{itemize}
		\item activation shift (see section \ref{shift}) that eliminates the ambiguity of zero entries in Gram matrices and improves results while accelerating convergence across different style images and style transfer methods;
		\item augmenting the style representation by using 16 layers (contrary to 5 in \cite{Gatys15}) and by considering a chain of inter-layer feature correlations (see sections \ref{cor}, \ref{chain} and \ref{blur});
		\item geometric weighting scheme to soften the style/content separation and to prioritize simple style features when repainting (see section \ref{weighting}).
	\end{itemize}

	These changes have consistently yielded improvements on the majority of style images in our experiments, ranging from marginal up to very significant, although at the cost of increased memory and computational cost compared to \cite{Gatys15}.
	
	Other suggestions discussed in section \ref{other} gave mixed results and may require further research in order for them to bare fruit.
	
	We believe our modifications to be most useful when using a style of ``average complexity''. 
	
	Simple, repetitive textures are very well transferred by the original algorithm in \cite{Gatys15}. However, it does fail often on more complex styles, e.g. images with a background and a foreground.
	
	On the other end of complexity range, in the context of photorealistic style transfer, \cite{CNNMRF} remains unchallenged. However, it does require a very good content match between content and style. Otherwise results may be inferior  to our approach, with violations of criterion 2 caused by certain content parts disappearing due to lack of distinct corresponding patches in the style image.
	
	However, as one can see in section \ref{exp}, our approach is still far from our goal of satisfying the two criteria in section \ref{intro}: uniform background sometimes gets fragmented into differently stylized regions (violating criterion 1), and the foreground object may sometimes blend with background (violating criterion 2). It is merely a step in this direction, that we hope will be useful for future research.

\section*{Acknowledgments}
 We would like to thank our professor Iasonas Kokkinos for helpful discussion and for providing hardware to run our experiments.

\newpage
\printbibliography

\end{document}